\documentclass[11pt]{article}
\usepackage{nodalida2023}
\usepackage{float}
\usepackage{times}
\usepackage{latexsym}
\usepackage{graphicx}
\usepackage{url}
\usepackage{hyperref}
\usepackage{amsmath}
\usepackage{comment}
\usepackage{makecell}
\usepackage{multirow}
\usepackage{subcaption}
\usepackage{caption}
\usepackage[title]{appendix}
\usepackage{tabularx}
\usepackage{booktabs}
\usepackage{longtable}
\usepackage{footnote}
\usepackage{longtable}

\usepackage{booktabs}
\usepackage{adjustbox}

\aclfinalcopy 
\title{\vspace*{-0.1cm}
\textsc{MaLei} at the PLABA Track of TREC 2024: RoBERTa for Term Replacement -- LLaMA3.1 and GPT-4o for Complete Abstract Adaptation }

\author{
         Zhidong Ling $^{1}$ Zihao Li $^{2}$ Pablo Romero $^{3}$ \\
         \textbf{Lifeng Han $^{2,4*}$ Goran Nenadic $^{2}$} \\
         $^1$ Tokyo Metropolitan University,
         $^2$ The University of Manchester\\
         $^3$ Manchester Metropolitan University, 
         $^4$ Leiden University\\
         {\tt ling-zhidong@ed.tmu.ac.jp}, 
         {\tt 22466459@stu.mmu.ac.uk}\\
         {\tt \{zihao.li-2,g.nenadic\}@manchester.ac.uk}, {\tt l.han@lumc.nl} \\ $^*$ corresponding author
}

\date{}

\begin{document}
\maketitle
\begin{abstract}

This report is the system description of the \textsc{MaLei} team (\textbf{Manchester} and \textbf{Leiden}) for the shared task Plain Language Adaptation of Biomedical Abstracts (PLABA) 2024 (we had an earlier name BeeManc following last year), affiliated with TREC2024 (33rd Text REtrieval Conference \url{https://ir.nist.gov/evalbase/conf/trec-2024}). 
This report contains two sections corresponding to the two sub-tasks in PLABA-2024. 
In task one (term replacement), we applied fine-tuned ReBERTa-Base models to identify and classify the difficult terms, jargon, and acronyms in the biomedical abstracts and reported the F1 score (Task 1A and 1B). 
In task two (complete abstract adaptation), we leveraged Llamma3.1-70B-Instruct and GPT-4o with the one-shot prompts to complete the abstract adaptation and reported the scores in BLEU, SARI, BERTScore, LENS, and SALSA.
From the official Evaluation from PLABA-2024 on Task 1A and 1B, our \textbf{much smaller fine-tuned RoBERTa-Base} model ranked 3rd and 2nd respectively on the two sub-tasks, and the \textbf{1st on averaged F1 scores across the two tasks} from 9 evaluated systems. Our LLaMA-3.1-70B-instructed model achieved the \textbf{highest Completeness} score for Task 2.
We share our source codes, fine-tuned models, and related resources at \url{https://github.com/HECTA-UoM/PLABA2024}

\end{abstract}

\section{Background}
Health literacy, or the ability of individuals to comprehend and apply health information for informed decision-making, is one of the central focuses of the Healthy People 2030 framework in the US. 
Even though biomedical information is highly accessible online, patients and caregivers often struggle with language barriers, even when the content is presented in their native language.

The shared task PLABA aims to harness advances in deep learning to empower the automatic simplification of complex scientific texts into language that is more understandable for patients and caregivers.
Despite substantial obstacles to effective implementation, the goal of the PLABA track is to improve health literacy by translating biomedical abstracts into plain language, making them more accessible and understandable to the general public \footnote{\url{https://bionlp.nlm.nih.gov/PLABA-2024/}}.
Following our participation on the PLABA-2023 shared task using large language models (LLMs) such as ChatGPT, BioGPT, and Flan-T5, and Control Mechanisms \cite{li-et-al-PLABA-2023-ieeeichi}, in this work, we introduce our system participation to the PLABA-2024. 
Instead of end-to-end biomedical abstract simplification as in PLABA-2023, in this year, PLABA-2024 introduced more granular-steps, including Term Replacement for Task-1 and Complete Abstract Adaption for Task-2, which we will describe in detail for our methodologies via fine-tuning RoBERTa-Base model for Task-1 and prompting LLMs (LLaMa-3.1-70B and GPT4o) for Task-2.

\section{PLABA-2024 Task 1: Term Replacement}

\subsection{Introduction for Task 1}
Task 1 does not require a full adaptation process. 
Instead, the system will identify challenging terminology, determine appropriate strategies for addressing these terms, and offer suitable replacements.
This task is split into three subtasks: 

\paragraph{1A: Identifying non-consumer terms}
This is a Name Entity Recognition (NER) task.
The objective is to to extract a list of exact phrases from the given abstracts, each representing a concept that a consumer would not understand.

\paragraph{1B: Classifying replacement}
This is a multi-class multi-label token classification task.
Systems should identify any and all types of replacements that would be appropriate for each identified non-consumer term.

\paragraph{1C: Generation}
Provide text for each positive label from 1B (except for the ``OMIT'' label). For this task, there aren't popular submissions, so we will skip this task.

\subsection{Methodology}

\begin{figure*}
    \centering
    \includegraphics[width=1\linewidth]{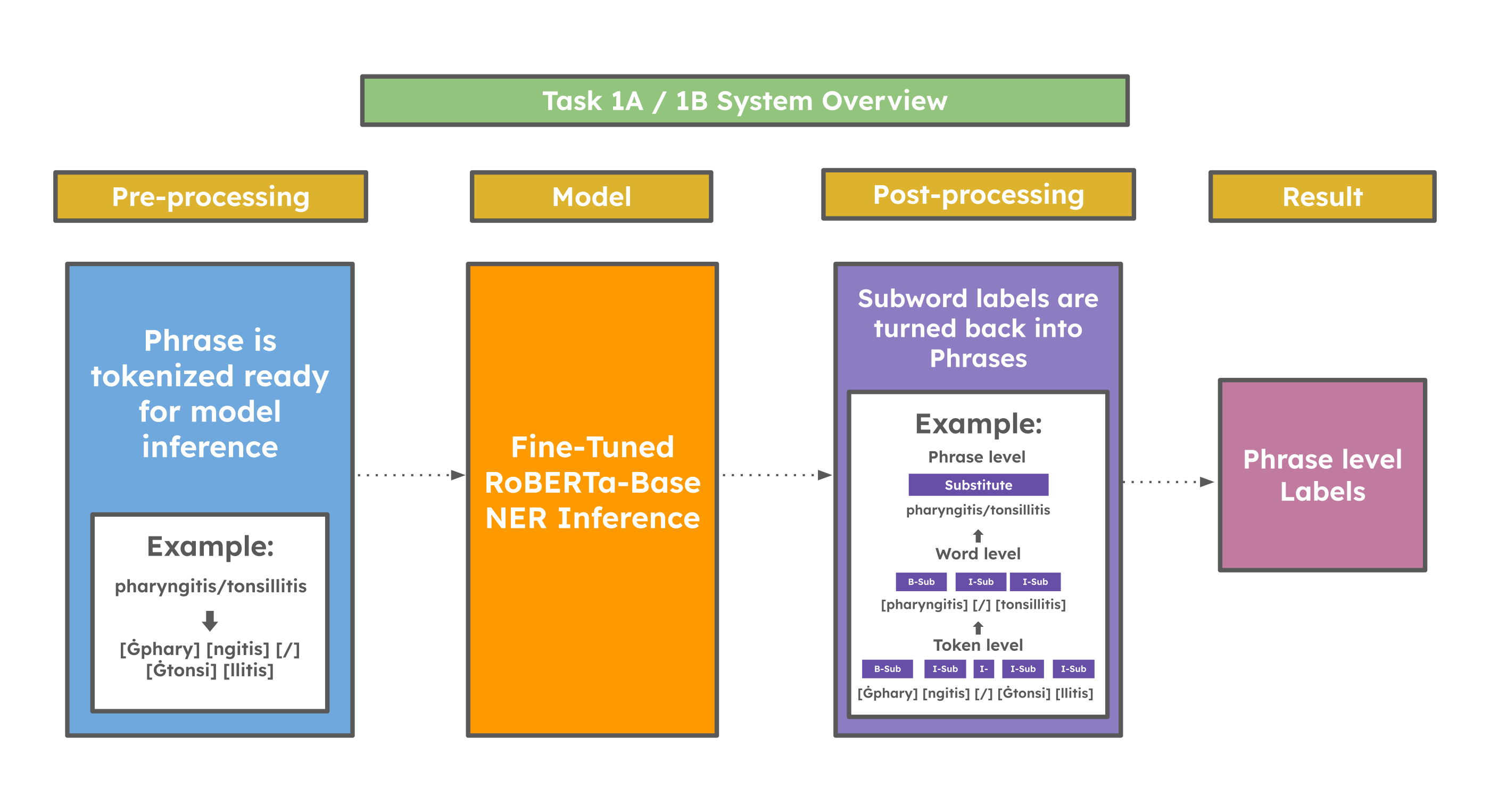}
    \caption{System overview for Task 1A and 1B}
    \label{fig:system_task1}
\end{figure*}

\subsubsection{Corpus}
\label{sec:corpus}
In this task, we leveraged the PLABA 
corpus~\cite{attal2023dataset}, which contains biomedical queries and the abstracts of top-ten papers corresponding to the query from PubMed. The sentences in each abstract are manually aligned and simplified by experts. In addition, the difficult phrases are tagged with proper labels indicating the replacement method. The original dataset is in JSON format and one sample is shown in Fig\ref{fig:training-data-example}.

There are five different kinds of replacement labels, and the task requires systems to assign these labels non-exclusively to the phrases.
\begin{itemize}
    \item Substituted: the term is jargon with a common alternative (e.g. "myocardial infarction" can be "heart attack").
    \item Explained: there is no alternative, or the term is important to the topic, and it should be explained. For example: "This study looked at treatments for sleep apnoea (when you stop breathing while sleeping)."
    \item Generalized: the term can be replaced with a more general category without losing its significance. For example: "Clearing of the infection is confirmed with a [\textit{Nucleic Acid Amplification Test (NAAT)}] common lab test."
    \item Exemplified: the term has a specific example that would give a general audience an idea of what it is. For example: "Depression is common in people with neurodegenerative diseases (like Parkinson’s)."
    \item Omitted: The term is not relevant to understanding the sentence or too technical to explain, and does not need to appear in a consumer version.
\end{itemize}

To adapt this data to a form that is suitable for an NER task, we need to map these labels to the phrases in the sentences.

\subsubsection{Data Preprocessing}
We introduce how we modify the training data to a format for our task setting.
For each sentence, terms that are difficult to read are given human-annotated replacement labels.
Terms can be a single word or a phrase, and label types can also be multiple.

For task 1A, we define it as a word-level binary NER task.
Because this subtask is covered by subtask 1B, we did not develop a system to specifically address this task.
The output for task 1A is the words that are assigned not the O label in task 1B.

\begin{figure}
    \centering
    \includegraphics[width=1\linewidth]{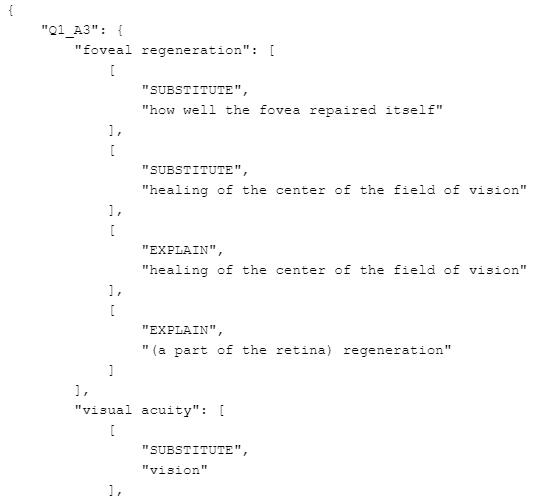}
    \caption{ Example of  training corpus for Task 1}
    \label{fig:training-data-example}
\end{figure}

For Task 1B, labels must be mapped to subword-level labels during model fine-tuning, as the model's tokenizer segments sentences into subword units.
The original training data shown in Figure \ref{fig:training-data-example} contains phrase-level annotations, which are not directly compatible with the subword-level classification required by our NER model. To address this issue, we implemented a two-step label transformation process.
First, we decomposed the annotated phrases into individual words, applying the BIO tagging scheme: the first word of each phrase receives a B-[tag] (Beginning), while subsequent words receive I-[tag] (Inside) labels.
Second, to accommodate the model's tokenizer, which further segments words into subwords, we developed the following labelling strategy:

\begin{itemize}
    \item For words originally labeled with B-[tag]: the first subword maintains the B-[tag], while subsequent subwords receive I-[tag] labels
    \item For words originally labeled with I-[tag]: all resulting subwords inherit the I-[tag] label
    \item Context words outside the annotated phrases are labelled with \textit{O} (Outside)
\end{itemize}

This approach ensures consistent label propagation from phrase-level annotations to subword-level classifications needed to train the NER model.

The classification uses the following label types:
\begin{itemize}
\item SUBSTITUTE
\item EXPLAIN
\item EXEMPLIFY
\item GENERALIZE
\item OMIT
\end{itemize}
Each type follows the BIO tagging scheme with Beginning (B-) and Inside (I-) variants, plus an additional Outside (O) label, yielding eleven labels in total.
Finally, we split the official PLABA-2024 data into training, validation, and test sets with a ratio of 8:1:1, resulting in 836 sentences for training, 105 for validation, and 105 for testing.
Table \ref{tab:label_counts} shows the distribution of the label for each set.

\begin{table}[h]
\centering
\small

{\renewcommand{\arraystretch}{1.3}  
\begin{tabular}{|c|r@{\hspace{10pt}}r@{\hspace{10pt}}r|}
\hline
\textbf{Label} & \textbf{Train} & \textbf{Valid} & \textbf{Test} \\ \hline
O              & 17703          & 2005           & 2411          \\
B-SUBSTITUTE   & 1989           & 264            & 239           \\
I-SUBSTITUTE   & 1385           & 181            & 191           \\
B-EXPLAIN      & 1047           & 160            & 122           \\
I-EXPLAIN      & 549            & 68             & 62            \\
B-GENERALIZE   & 317            & 50             & 39            \\
I-GENERALIZE   & 419            & 86             & 52            \\
B-EXEMPLIFY    & 44             & 6              & 6             \\
I-EXEMPLIFY    & 37             & 2              & 7             \\
B-OMIT         & 309            & 45             & 50            \\
I-OMIT         & 314            & 38             & 79            \\ \hline
\end{tabular}}  
\caption{Label Counts for Training, Validation, and Test Sets for Task 1B: 11 labels}
\label{tab:label_counts}
\end{table}

\subsubsection{Model Description}

For this multi-label, multi-class classification task, we selected the Robustly Optimized BERT Pretraining Approach (RoBERTa)~\cite{liu-roberta-2019} as our model of choice. 
RoBERTa is a transformer-based language model.
It enhances the pretraining methodology of Bidirectional Encoder Representations from Transformers (BERT) to achieve superior performance in natural language processing tasks. 
Specifically, RoBERTa removes the next-sentence prediction objective, increases the amount of training data, and employs larger batch sizes and learning rates during training. 
These optimizations enable RoBERTa to outperform BERT on several benchmarks, demonstrating its advanced capabilities in understanding and generating human language.
We used the open-source code from \textsc{INSIGHT}Buddy-AI for RoBERTa fine-tuning \cite{romero2024insightbuddyaimedicationextractionentity}.

\subsection{Experiments}

\begin{figure*}[!h]
    \centering
    \includegraphics[width=1\linewidth]{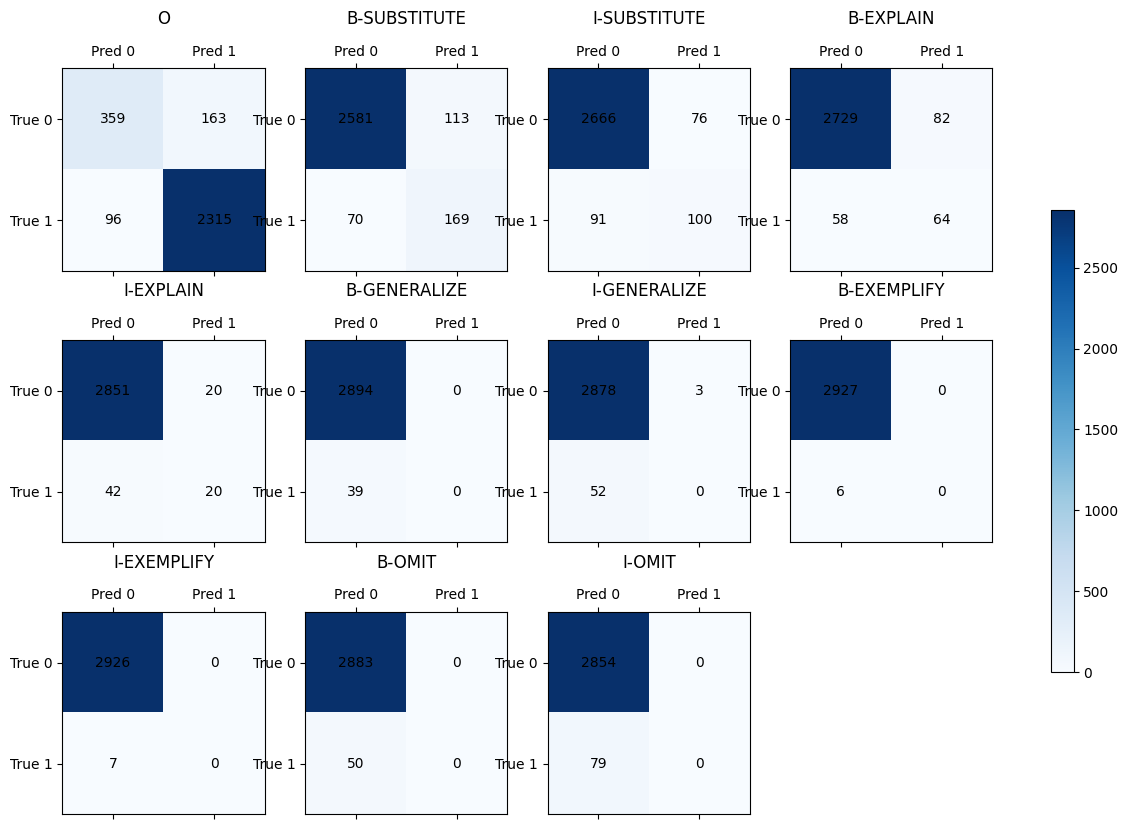}
    \caption{The confusion matrices for the results on the test set of Task-1B for each label category.}
    \label{fig:confusion-matrices}
\end{figure*}

\begin{figure*}[!h]
    \centering
    \includegraphics[width=1\linewidth]{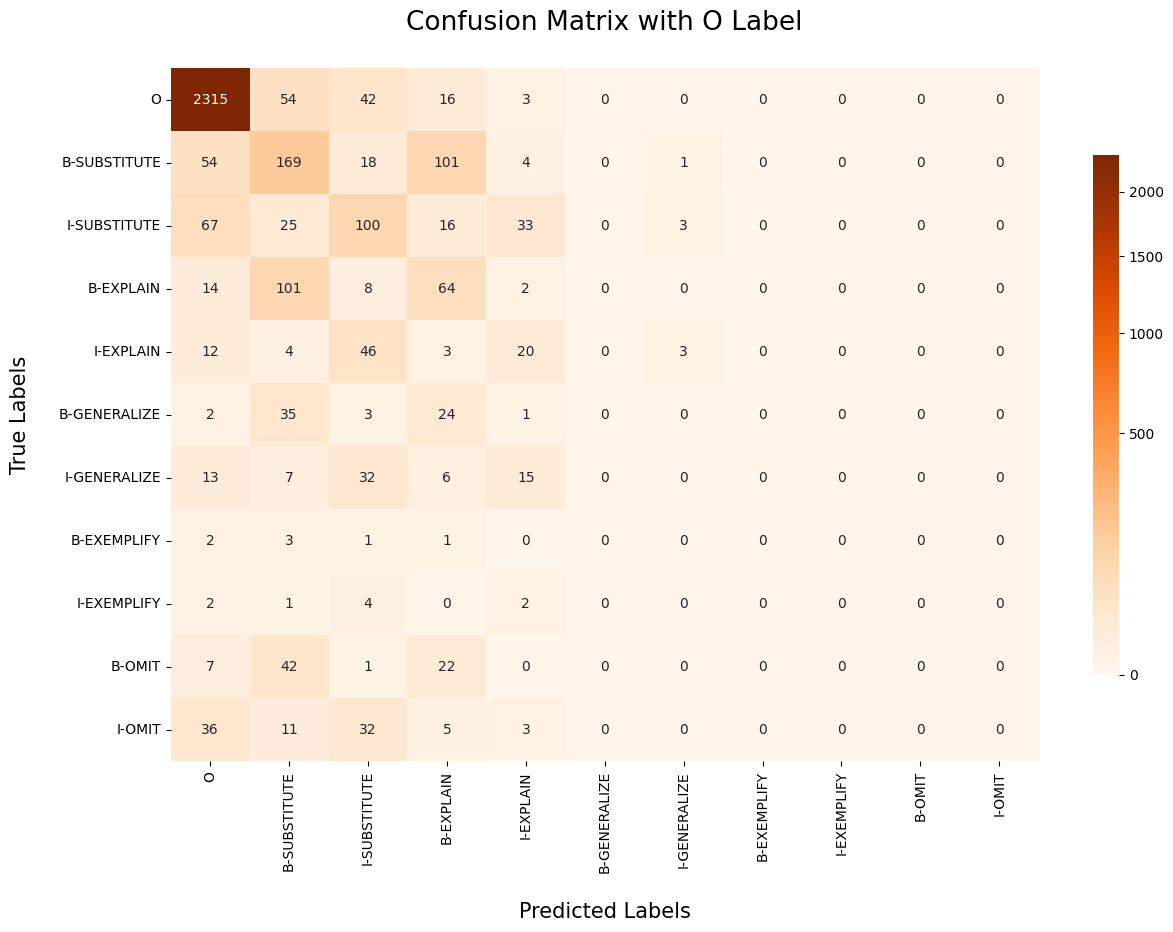}
    \caption{The confusion matrix between all labels on the test set (our 8:1:1 split) for Task 1B. }
    \label{fig:confusion-matrices-in-one}
\end{figure*}

\paragraph{Fine tuning process}

We add a linear layer after RoBERTa's final hidden layer. 
The system converts the output for each token into a set of probabilities for the 11 classes using the sigmoid function. 
We use Binary Cross Entropy as our loss function, specifically implementing it through PyTorch's BCEWithLogitsLoss\footnote{\url{https://pytorch.org/docs/stable/generated/torch.nn.BCEWithLogitsLoss.html\#bcewithlogitsloss}}. 
A threshold of 0.5 is applied after the sigmoid function to determine if a token belongs to a particular label. 
Because we use sigmoid rather than softmax, each token can be assigned multiple labels. 
The detailed fine-tuning settings are shown in Table \ref{tab:fintuning_args}.

\paragraph{Post-processing}
After the system assigns subword labels, we merge them into word-level labels.
Our strategy for combining subwords into words follows these rules:
\begin{enumerate}
    \item A subword token beginning with the special character ``Ġ" indicates the start of a new word.\footnote{``Ġ" is the special token from RoBERTa tokenizer; it is automatically attached to a subword when we do ``tokenizer.decode'' function.}
    \item If the subsequent token also begins with ``Ġ", the current token is treated as a complete word and retains its label.
    \item When a token with a leading ``Ġ" is followed by one or more tokens without ``Ġ", these tokens are merged into a single word. The merged word inherits the label from the initial ``Ġ" token. For example, the tokens [`Ġdoc', `us', `ate'] with labels ['B-SUBSTITUTE', 'B-EXPLAIN'], ['I-SUBSTITUTE', 'I-EXPLAIN'], ['I-SUBSTITUTE', 'I-EXPLAIN'] merge to form 'docusate' with the labels ['B-SUBSTITUTE', 'B-EXPLAIN'].
    \item If a word's first token is labelled 'O' but its later subwords have other labels, the whole word takes the first non-'O' label. For example: ['Ġdoc'('O'), 'us'('O'), 'ate'('I-SUBSTITUTE')] becomes 'docusate'('I-SUBSTITUTE').
    \item Special sequence tokens (\textless s \textgreater, \textless /s \textgreater) added by RoBERTa's tokenizer are removed as they are not part of the actual text content.
\end{enumerate}

As the output of the format is phrase level, we need to merge the words back into phrases, here is our merging strategy for forming phrases:

\begin{enumerate}
    \item When a token has a B-[tag], we combine it with all following I-[tag] tokens to form a phrase. For example: ['repeat'(B-SUBSTITUTE), 'infect'(I-SUBSTITUTE), 'ion'(I-SUBSTITUTE), 'rates'(I-SUBSTITUTE)] forms the phrase "repeat infection rates" with label SUBSTITUTE.
    
    \item If we find an I-[tag] without a preceding B-[tag], we trace back to the first token starting with "Ġ" to mark the phrase beginning.
\end{enumerate}
The output consists of identified difficult phrases and their corresponding replacement labels at the phrase level.

\subsection{Results/Discussion}

We report Automatic Evaluation at the word level for task 1B in table \ref{tab:task1b-result}.
\begin{table*}[t]
    \centering
    \begin{tabular}{c|ccc}
        \hline
         & F1-micro & Precision-micro & Recall-micro\\\hline
        word level(w/ ['O'] label) & 0.836 & 0.858 & 0.819\\
        word level(w/o ['O'] label) & 0.413 & 0.546 & 0.417\\\hline
        & F1-macro & Precision-macro & Recall-macro\\\hline
        word level(w/ ['O'] label) & 0.274 & 0.276 & 0.276\\
        word level(w/o ['O'] label) & 0.206 & 0.211 & 0.208\\\hline
        
    \end{tabular}
    \caption{System performance on the test set (our 8:1:1 split of \cite{attal2023dataset}) for task 1B.}
    \label{tab:task1b-result}
\end{table*}

The RoBERTa-base model's performance shows significant limitations. While the word-level NER achieves a micro-F1 score of 0.836 when including ['O'] labels, this score drops substantially to 0.413 when excluding them. The confusion matrices for all 11 labels are shown in Fig \ref{fig:confusion-matrices}.
This performance degradation primarily stems from the model's poor prediction of low-frequency labels in the training data, particularly EXEMPLIFY, OMIT, and GENERALIZE. The limited training instances for these labels appear insufficient for effective model fine-tuning. Analysis of the training process reveals that the model rarely correctly classified these minority labels, suggesting room for improvement in handling class imbalance.
The data and label distribution in Table \ref{tab:label_counts} also reveals that the lower macro score is due to the zero accuracy on the low-ratio labels (Figure \ref{fig:confusion-matrices-in-one} lower part of the diagonal line). Even though these labels have a small appearance in the entire dataset, macro score assigns each label class the same weight, bringing down the overall macro scores.
However, the overall performance of the system is still reasonable instead of nonsense, which can be seen in the upper part of Table \ref{tab:task1b-result}, where the micro F1 score without the O label is still 0.413 on all special labels. 

\section{PLABA-2024 Task 2: Complete Abstract Adaptation}

\subsection{Introduction}
Task 2 involves the end-to-end adaptation of biomedical abstracts for a general audience by utilizing plain language. 
Given a collection of abstracts as the source material, the system will generate a corresponding plain language adaptation as the output for each sentence within the source text.

\subsection{Methodology}

To generate simplified versions of abstract sentences, we employed the Llama-3.1-70B~\cite{dubey2024llama3herdmodels} instruction tuned version \footnote{\url{https://huggingface.co/meta-llama/Llama-3.1-70B-Instruct}} and GPT-4o~\cite{openai2024gpt4technicalreport} with one-shot prompt only.

\subsubsection{Corpus}

In this shared task, we stick with the same PLABA corpus~\cite{attal2023dataset}, which is introduced in section \ref{sec:corpus}, but it is mostly used for evaluation only. In the corpus, there are 40 abstracts containing 7604 sentences in total with a reference or references.

\subsubsection{Metrics}

In this shared task, we mainly rely on automatic evaluation metrics, which can be classified into reference-based and reference-less metrics. In reference-based metrics, BLEU~\cite{papineni-etal-2002-bleu}, SARI~\cite{xu-etal-2016-sari}, BERTScore~\cite{bert-score} and LENS~\cite{maddela-etal-2023-lens} are chosen due to their popularity in text simplification areas. SALSA~\cite{heineman-etal-2023-dancing}, which is a new-developed metric, is used for reference-less evaluation.

\paragraph{BLEU}
The Bilingual Evaluation Understudy (BLEU) score~\cite{papineni-etal-2002-bleu} is a metric widely used in machine translation to evaluate the quality of the generated text. As text simplification can be regarded as monolingual text translation, it is also accepted and used in the early stages of text simplification studies. It measures the n-gram (from 1 to 4) text similarity between the generated text and the single or multiple references.

\paragraph{SARI}
The System Output Against Reference and Input (SARI) score~\cite{xu-etal-2016-sari} is a metric designed to evaluate text simplification tasks. Similar to BLEU, SARI also measures the n-gram similarity between the output and references. Additionally, the score is determined based on three operations: adding, deleting, and retaining words from the original sentence. Each operation is assessed by comparing it to one or more reference simplified texts to measure how closely the simplified text matches expert simplifications. As a result, it better aligns with human evaluations in text simplification. simplification.

\paragraph{BERTScore}
Unlike the two above-mentioned metrics, BERTSCore~\cite{bert-score} measures the semantic similarity between the output and reference instead of n-gram similarity. By leveraging the deep contextualized embeddings from BERT~\cite{devlin-etal-2019-bert}, it can measure the semantic similarity among the tokens. Then it calculates the sentence similarity by token matching and maximising the matrix between output and references.

\paragraph{LENS}
Learned Evaluation of Non-Canonical Sequences (LENS)~\cite{maddela-etal-2023-lens} is yet another metric for evaluating text generation tasks. The embedded LENS machine learning model in LENS makes it different from the previous metrics. It evaluates the generated text based on semantic and contextual similarity, which improves the adaptability and accuracy in tasks where the generated text may differ in surface form from reference text but still be semantically correct.

\paragraph{SALSA}
\textbf{S}uccess and F\textbf{A}ilure driven \textbf{L}inguistic \textbf{S}implification \textbf{A}nnotation (SALSA)~\cite{heineman-etal-2023-dancing} is an edit-based human annotation framework. By training LENS~\cite{maddela-etal-2023-lens} on 19 thousand edit annotations gathered through SALSA, the authors developed LENS-SALSA, which we refer to simply as SALSA in this report. The main difference between SALSA and LENS is that SALSA can measure performance based on self-predicted references, which is beneficial for evaluating the test set.

\subsection{Experiment}

\subsubsection{prompting process}

To generate simplified sentences in the abstract, we started with a simple 1-shot prompt that has one example in the prompt. Then based on the output of test cases, we kept asking the model to self-adjust the prompt until it almost generated the expected outcome.
We list the prompts we used in Tables \ref{tab:promptgpt4o} and \ref{tab:promptllama} in the appendix. 

\subsubsection{post-processing}

Sometimes models will output some pre-words such as "Here's a simplified version: " and "**Simplified**:", we checked the outputs and used the rule to delete these pre-words.
Some outputs have detailed explanations for how and why did I simplify this sentence.
For this kind of output, we used regular expressions to extract the simplified sentence as possible.

\subsection{Results and Discussion}

\paragraph{Automatic Evaluation}

\begin{table}[th!]
    \centering
    \begin{tabularx}{\columnwidth}{p{2cm}|XXXX}
        Model & SALSA & LENS & SARI & BERT(F$_1$)  \\\hline
        
        Llama 3.1 instruct-70B & 73.89 & 52.79 & 36.37  &0.88  \\
        
        GPT-4o & 74.28 & 64.75  & 37.72 &0.92  \\
    \end{tabularx}
    \caption{System performance on the full training set across SALSA, LENS, SARI and BERT for BERTScore. The training set is provided by PLABA dataset~\cite{attal2023dataset}}
    \label{tab:eval_res}
\end{table}

\begin{table}[th!]  
    \centering
    \begin{tabularx}{\columnwidth}{p{2cm}|X}
        Model & SALSA \\
        \hline
        Llama 3.1 instruct-70B &  61.79 \(\pm\) 0.30\\
        GPT-4o & 73.08 \(\pm\) 0.24 \\
    \end{tabularx}
    \caption{System performance on the test set under the SALSA (95\% confidence interval). The test set is provided by PLABA-2024, and is different from the training set.}
    \label{tab:test_res}
\end{table}

Tables \ref{tab:eval_res} and \ref{tab:test_res} show the performance difference between the two models across the metrics. In Table \ref{tab:eval_res}, all four metrics consistently favour GPT-4o over Llama and show some credibility of SALSA as a reference-less metric. Similarly in Table \ref{tab:test_res}, the two models show an aligned performance gap, compared to Table \ref{tab:eval_res}.
Comparing these two tables (\ref{tab:eval_res} and \ref{tab:test_res}): 
\begin{itemize}
    \item With (source, output, reference) set, the difference of LLaMa-3.1-70B and GPT-4o on different metrics is not that large margin, except for LENS, as in Table \ref{tab:eval_res}.
    \item With (source, output) no reference: the SALSA score gap increased dramatically, as in Table \ref{tab:test_res}.
\end{itemize}

\section{Official Ranking from PLABA-2024}
\subsection{Task-1 (A and B)}
Table \ref{tab:plaba24-official-1ab} shows the system ranking from PLABA-2024 official organisers on Task 1A and 1B using F1 scores, 1C being not popular among the submissions.
There are overall 9 systems evaluated by PLABA-2024 \cite{plaba2024-nist}.
Out of the 9 systems, our team UM ranks the third for Task 1A, and the 2nd for Task 1B via F1 scores. There are a few intersting findings from the results.

\begin{itemize}
    \item Firstly, it is worth to note that \textit{for both tasks, our model scores are very close to the highest system}, e.g. for Task 1A, our F1 score is 0.4787 using Roberta-Base, after the first two systems 0.5036 from ReBERTa-GBC and 0.4885 from Gemini-1.5. For Task 1B, our F1 score 0.7765 is even closer to the top-1 score 0.7788 from MLPClassifier system. In general, our fine-tuned system is almost as good as the top-1 system at identifying the complicated term, but we are much better at classifying the next to-do step in the term processing (11 labels = 1 +  5x2) compared to the top-1 system for Task-1A (0.7765 vs 0.6838) winning almost 10 absolute point difference. It is also interesting to see that the Top-1 system for Task-1B (F1 0.7788) from MLPClassifier, produced the lowest F1 score for Task-1A, i.e. 0.0459, which means it actually cannot identify most of the complicated terms (only made into 5\%).
    \item Secondly, our \textit{averaged F1 score across Task 1A and 1B} is the \textbf{highest} among all 9 systems, 0.6276=(0.4787+0.7765)/2. It is also the only system whose Averaged F1 score is above 0.60.
    \item Thirdly, It is interesting to see that \textbf{much smaller models can outperform the extra-large LLMs} on Task 1A and 1B, e.g. our RoBERTa-base fine-tuned wins Mistral, GPT, Gemini-1.5 on these two tasks. This is related to the findings from \cite{li-et-al-PLABA-2023-ieeeichi} for the PLABA-2023 shared task.
\end{itemize}


\begin{table*}[h!]
    \centering
    \begin{adjustbox}{width=\textwidth}
    \begin{tabular}{llcccc}
        \toprule
        \textbf{Team} & \textbf{Run} & \textbf{Rank} & \textbf{Task 1A (F1)} & \textbf{Task 1B (F1)} & \textbf{Average Score} \\
        \midrule
        BU & MLPClassifier-identify-classify-replace-v1 & 1 & 0.0459 & \textbf{0.7788} & 0.4124 \\
        CLAC & mistral & 1 & 0.4410 & 0.6663 & 0.5537 \\
        CLAC & gpt & 2 & 0.3767 & 0.3795 & 0.3781 \\
        IITH & First & 1 & 0.1956 & 0.7014 & 0.4485 \\
        \underline{UM} & \underline{Roberta-base} & 1 & \textit{0.4787} & \textit{0.7765} & \textbf{0.6276} \\
        ntu\_nlp & gemini-1.5-pro\_demon5\_replace-demon5 & 1 & \textit{0.4885} & 0.6335 & 0.5610 \\
        ntu\_nlp & gemini-1.5-flash\_demon5\_replace-demon5 & 2 & 0.4431 & 0.6544 & 0.5488 \\
        ntu\_nlp & gpt-4o-mini\_demon5\_replace-demon5 & 3 & 0.4518 & 0.6197 & 0.5358 \\
        Yseop & roberta-gbc & 1 & \textbf{0.5036} & 0.6838 & 0.5937 \\
        \bottomrule
    \end{tabular}
    \end{adjustbox}
        \caption{Task 1A and Task 1B (F1) Results: Roberta-base highest Averaged-Score\cite{plaba2024-nist}}
        \label{tab:plaba24-official-1ab}
\end{table*}

\subsection{Task-2}
For Task 2, even though GPT4o outscored LLaMa-3.1-70B in our model development phase,
to understand better how open-science LLMs perform on biomedical text simplification tasks,
we selected LLaMa-3.1-70B as our prioritised output over the GPT4o-instruct for official human evaluation in the PLABA-2024.
As shown in Table \ref{tab:PLABA-2024-official-task2}, 
it produced interesting outcomes with the \textbf{highest} Completeness score (0.9481) vs the \textit{lowest} Accuracy score (0.6088). 

According to the human evaluation guidelines from PLABA-2024, Accuracy is ``Outputs should contain the accurate information'', while Completeness is ``Outputs should seek to minimize information lost from the original text''.
So, LLaMa-3.1-70B-instructed can do its best to avoid information loss among the 8 evaluated systems, but it did not produce the most accurate outputs. These are somehow contradicting findings but it might be because the model produced some misleading outputs in addition to keeping the source information. We need further qualitative analysis on this assumption.

\begin{table*}[h!]
    \centering
    \begin{adjustbox}{width=\textwidth}
    \begin{tabular}{lccccc}
        \toprule
        \textbf{Run} & \textbf{Accuracy} & \textbf{Completeness} & \textbf{Simplicity} & \textbf{Brevity} & \textbf{Final avg.} \\
        \midrule
        LLaMA-8B-4bit-MedicalAbstract-seq-to-seq-v1 & 0.8545 & 0.9180 & 0.6533 & 0.3978 & 0.7059 \\
        \underline{LLaMa 3.1 70B instruction 2nd run} & 0.6088 & \textbf{0.9481} & 0.5561 & 0.5004 & 0.6533 \\
        TREC2024\_SIB\_run3 & 0.7506 & 0.7716 & 0.7997 & 0.6484 & 0.7426 \\
        UAms-ConBART-Cochrane & 0.9534 & 0.9398 & 0.6851 & 0.6171 & 0.7988 \\
        gpt35\_dspy & 0.9167 & 0.9386 & 0.6974 & 0.5478 & 0.7751 \\
        mistral-FINAL & 0.6694 & 0.7552 & 0.5149 & 0.2875 & 0.5567 \\
        plaba\_um\_fhs\_sub1 & 0.9067 & 0.9138 & 0.7814 & 0.6799 & 0.8205 \\
        task2\_moa\_tier1\_post & 0.8982 & 0.9444 & 0.7246 & 0.5284 & 0.7739 \\
        \bottomrule
    \end{tabular}
    \end{adjustbox}
    \caption{PLABA-2024 Task 2 results: LLaMa-3.1 with the highest Completeness \cite{plaba2024-nist}}
    \label{tab:PLABA-2024-official-task2}
\end{table*}

\section{Conclusion and Future Work}
In this report, we present the two systems submitted by the \textsc{MaLei} team for two different tasks in PLABA-2024. 
For Task 1, we submitted a classification system based on Roberta for Task 1a and 1b. 
The system's performance at the word level was not outstanding, highlighting the limitations of simple classification models in multi-label, multi-class tasks. 
For Task 2, we submitted end-to-end generation results based on LLaMa 3.1 and GPT-4o for the generation task. 
On the Evaluation set, GPT-4o performed slightly better than LLaMa. 
In the unsupervised evaluation on the test set, GPT-4o also outperformed LLaMa 3.1. 
Our findings demonstrate the potential of open-source models in text simplification tasks within the medical domain.

Regarding LLaMa-3.1-70B-instructed model producing the highest Completeness score but the lowest Accuracy according to the official human evaluation, we plan to carry out further qualitative analysis on its output to see if the system output contains misleading or hallucinated information in addition to the secured source knowledge. We can also use some specific human-in-the-loop metrics such as HOPE \cite{gladkoff-han-2022-hope} to categorise the error types.

\section{Acknowledgment}
We thank Nicolo and Sam for their valuable discussion and feedback during this project with their expertise from the experience of PLABA-2023.
The authors wrote the initial draft, used Generative AI to revise/polish it, then the authors reviewed the generated content. The authors used ChatGPT to refine the paper writing.
LH and GN were supported by the grant “Integrating hospital outpatient letters into the healthcare data space” (EP/V047949/1; funder: UKRI/EPSRC).
LH is also grateful to the EU grant 4D Pictures \url{https://4dpicture.eu/}.

\bibliographystyle{acl_natbib}
\bibliography{nodalida2023}


\onecolumn
\newpage

\begin{appendices}

\clearpage
\begin{table}[h]
\centering
\begin{tabular}{l|l}
\hline
\textbf{Argument}                      & \textbf{Value}                                \\ \hline
evaluation\_strategy                   & \texttt{epoch}                                \\ 
num\_train\_epochs                     & 10                                            \\ 
weight\_decay                          & 0.01                                          \\ 
save\_strategy                         & \texttt{epoch}                                \\ 
learning\_rate                         & 0.00005                                       \\ 
per\_device\_train\_batch\_size        & 32                                            \\ 
per\_device\_eval\_batch\_size         & 32                                            \\ 
dataloader\_num\_workers               & 4                                             \\ 
dataloader\_pin\_memory                & \texttt{True}                                 \\ 
optim                                  & \texttt{adamw\_torch}                         \\ 
fp16                                   & \texttt{True}                                 \\ 
warmup\_ratio                          & 0.1                                           \\ \hline
\end{tabular}
\caption{Fine-tuning arguments for model training for RoBERTa-base}
\label{tab:fintuning_args}
\end{table}

\begin{longtable}{| m{2.5cm} | m{12cm} |}
    \hline
    \textbf{Section} & \textbf{Content} \\
    \hline
    \endfirsthead

    \hline
    \textbf{Section} & \textbf{Content} \\
    \hline
    \endhead

    \hline
    \multicolumn{2}{r}{\textit{Continued on next page}} \\
    \hline
    \endfoot

    \hline
    \endlastfoot

    \textbf{Objective} & You are tasked with simplifying a complex medical text to make it easily understandable by non-medical professionals, such as patients or caregivers. \\
    \hline

    \textbf{Guidelines} &
    \begin{itemize}
        \item \textbf{Maintain Accuracy:} Ensure that the simplified text accurately reflects the original meaning. Do not omit critical details or introduce incorrect information.
        \item \textbf{Use Plain Language:} Replace medical jargon and complex terms with simple, everyday language. If a technical term is necessary, provide a brief explanation or analogy.
        \item \textbf{Shorter Sentences:} Break down long, complicated sentences into shorter, more manageable ones. Multiple sentences are allowed but must fit within a single line per numbered point.
        \item \textbf{Use Active Voice:} Where possible, use the active voice to make the text more direct and easier to read.
        \item \textbf{Provide Examples:} Use examples or analogies to help explain complex concepts, making them relatable to everyday experiences.
        \item \textbf{Simplify Structure:} Organize the information in a logical, easy-to-follow structure. Use bullet points or lists to highlight key points when applicable.
    \end{itemize} \\
    \hline

    \textbf{Output Guidelines} &
    \begin{enumerate}
        \item Align the output strictly with the numbered points from the original text.
        \item Maintain the numbering format exactly as it appears in the input.
        \item Each numbered point must be a single line but can include multiple sentences if necessary for clarity.
        \item Ensure that each simplified point corresponds directly to its numbered counterpart in the original text.
    \end{enumerate} \\
    \hline

    \textbf{Original Text Example} &
    \begin{enumerate}
        \item The dystonias are a group of disorders characterized by excessive involuntary muscle contractions leading to abnormal postures and/or repetitive movements.
        \item A careful assessment of the clinical manifestations is helpful for identifying syndromic patterns that focus diagnostic testing on potential causes.
        \item If a cause is identified, specific etiology-based treatments may be available.
        \item In most cases, a specific cause cannot be identified, and treatments are based on symptoms.
        \item Treatment options include counseling, education, oral medications, botulinum toxin injections, and several surgical procedures.
        \item A substantial reduction in symptoms and improved quality of life is achieved in most patients by combining these options.
    \end{enumerate} \\
    \hline

    \textbf{Simplified Text Example} &
    \begin{enumerate}
        \item Dystonias are disorders with a lot of uncontrollable muscle contractions. This causes awkward postures and/or repetitive movements.
        \item Checking the symptoms helps spot patterns and guides testing to find possible causes.
        \item If a cause is found, treatments specific to that cause may be available.
        \item When no specific cause is found, treatments focus on managing symptoms.
        \item Treatment options include counseling, education, oral medications, botox (a muscle relaxant), and surgeries.
        \item Combining these treatments usually leads to a noticeable decrease in symptoms and a better quality of life for most patients.
    \end{enumerate} \\
    \hline
\caption{Prompt used for GPT4o}
\label{tab:promptgpt4o}

\end{longtable}

\begin{longtable}{| m{2.5cm} | m{12cm} |}
    \hline
    \textbf{Section} & \textbf{Content} \\
    \hline
    \endfirsthead

    \hline
    \textbf{Section} & \textbf{Content} \\
    \hline
    \endhead

    \hline
    \multicolumn{2}{r}{\textit{Continued on next page}} \\
    \hline
    \endfoot

    \hline
    \endlastfoot

    \textbf{Objective} & Simplify the provided text by:
    1. Rephrasing complex sentences for clarity.
    2. Replacing or defining rarely-used terms. \\
    \hline

    \textbf{Guidelines} &
     - For sentences that seem complex, rephrase them in simpler terms.
     - If you encounter unfamiliar or rare words, either replace them with a commonly known synonym or provide a concise definition.

    Note: In the training samples, complex sentences are flagged with ` \textless  rephrase \textgreater` and rare terms with ` \textless  rare \textgreater`. However, these tokens won't appear in testing samples. You'll need to recognize and address such complexities independently.
    Examples:\\
    \hline
    
    \textbf{Original Text Example} &
    1.Original:
    \begin{enumerate}
        \item  Furthermore, the circumference of thighs was measured to assess the  \textless rare \textgreater postoperative swelling \textless  rare \textgreater.
        \item  A total of 444 hypertensive patients, aged between 27 to 65 years, without any recent hypertensive treatment, were included.
        \item   \textless rephrase \textgreater The tongue often obstructs the upper respiratory tract, especially in comatose patients or those with cardiopulmonary arrest. \textless rephrase \textgreater
    \end{enumerate} \\
    \hline

    \textbf{Simplified Text Example} &
    \begin{enumerate}
        \item Additionally, we measured thigh sizes to check for swelling after surgery.
        \item 444 patients, aged 27-65 with high blood pressure and no recent treatment, were studied.
        \item The tongue can block breathing, mostly seen in unconscious people or those who've had a sudden heart stoppage.
    \end{enumerate} \\
    \hline
\caption{Prompt used for LLaMa-3.1-70B instruct}
\label{tab:promptllama}

\end{longtable}
\end{appendices}

\end{document}